\documentclass[conference]{IEEEtran}
\IEEEoverridecommandlockouts
\usepackage{cite}
\usepackage{amsmath,amssymb,amsfonts}
\usepackage{textcomp}
\def\BibTeX{{\rm B\kern-.05em{\sc i\kern-.025em b}\kern-.08em
    T\kern-.1667em\lower.7ex\hbox{E}\kern-.125emX}}

\usepackage{balance} 

\usepackage{algorithmic}
\usepackage{graphicx}
\usepackage{textcomp}
\usepackage{xcolor}
\usepackage{verbatim}
\usepackage{hyperref}
\usepackage{booktabs}
\usepackage[autostyle]{csquotes}
\usepackage{afterpage}
\usepackage{caption}
\usepackage{float}
\usepackage{etoolbox}
\usepackage[]{algorithm2e}
\usepackage{censor}

\usepackage{enumitem}

\begin{document}

\title{Gamified and Self-Adaptive Applications for the Common Good: Research Challenges Ahead}

\author{
\IEEEauthorblockN{Antonio Bucchiarone\IEEEauthorrefmark{1}, Antonio Cicchetti\IEEEauthorrefmark{2}, Nelly Bencomo\IEEEauthorrefmark{3}, Enrica Loria\IEEEauthorrefmark{1}, Annapaola Marconi\IEEEauthorrefmark{1}}
\IEEEauthorblockA{\IEEEauthorrefmark{1}Fondazione Bruno Kessler, Trento, Italy \emph{\{bucchiarone,eloria,marconi\}@fbk.eu}}
\IEEEauthorblockA{\IEEEauthorrefmark{2}IDT Department, M{\"a}lardalen University, V{\"a}ster{\aa}s, Sweden \emph{antonio.cicchetti@mdh.se}}
\IEEEauthorblockA{\IEEEauthorrefmark{3}SEA, CS, Aston  University, Birmigham, UK \emph{nelly@acm.org}}
}

\maketitle

\begin{abstract}
Motivational digital systems offer capabilities to engage and motivate end-users to foster behavioral changes towards a common goal. In general these systems use gamification principles in non-games contexts. Over the years, gamification has gained consensus among researchers and practitioners as a tool to motivate people to perform activities with the ultimate goal of promoting behavioural change, or engaging the users to perform activities that can offer relevant benefits but which can be seen as unrewarding and even tedious.

There exists a plethora of heterogeneous application scenarios towards reaching the common good that can benefit from gamification.
However, an open problem is how to effectively combine multiple motivational campaigns to maximise the degree of participation without exposing the system to counterproductive behaviours. 

We conceive motivational digital systems as multi-agent systems: self-adaptation is a feature of the overall system, while individual agents may self-adapt in order to leverage other agents’ resources, functionalities and capabilities to perform tasks more efficiently and effectively. Consequently, multiple campaigns can be run and adapted to reach common good. At the same time, agents are grouped into micro-communities in which agents contribute with their own social capital and leverage others' capabilities to balance their weaknesses.

In this paper we propose our vision on how the principles at the base of the autonomous and multi-agent systems can be exploited to design multi-challenge motivational systems to engage smart communities towards common goals. We present an initial version of a general framework based on the MAPE-K loop and a set of research challenges that characterise our research roadmap for the implementation of our vision.
\end{abstract}

\begin{IEEEkeywords}
Gamification, Multi-Agent Systems, Self-Adaptation, Societal Challenges
\end{IEEEkeywords}

\section{Motivation}

``How can we engage a person X in joining the cause Y, which may be perceived as time consuming, tedious, or out of X's interests? How can we locate and build-up appropriate (micro-)communities to maximise the engagement to cause Y by members of that community? What methods can we adopt to recognise and avoid behavioural tricks from playing against the system through unwanted interactions? How do we trade-off people's behaviours so to accommodate individuals' interests and abilities while still achieving competing causes, within the same environment?''

Those simple questions underpin the bold general objective of our vision: advanced  technologies leveraging social networking and its impacts for enabling communities to \textbf{collaboratively reach the common good} by \textit{sharing} people's strengths while \textit{improving} their weaknesses related to their \textbf{social capital} \cite{morrison_2013,Ostrom2000,PetruzziPB17}. Social capital, in this paper, is a set of characteristics that describes each individual’s capability and contribution to solve specific \textit{societal challenges} (i.e. improve inclusion, reduce environmental impact, have healthy habits)\cite{RR-479-EC}. Societal challenges can, ideally, be aligned to the Sustainable Development Goals set by the UN \footnote{\url{https://sustainabledevelopment.un.org/post2015/transformingourworld}} (e.g., climate action, sustainable cities and communities, and good health and well-being). 
 
We consider striving for the common good as a way to empower each individual, by taking advantage of potentials of collaborations while creating awareness of antagonistic interests. Some antagonistic interests may arrive to unwanted effects that may need to be removed at least up to a point \cite{morrison_2013,BucchiaroneDPCS20}.

\textbf{Gamification mechanisms} have the general merit of stimulating behavioural changes in a lightweight manner, that is by involving people in game-like scenarios where a certain task accomplishment is rewarded with a virtual or physical prize \cite{HervasRMB17}. 
Gamification is highly more effective when \textbf{personalisation} and \textbf{adaptivity} are included. Adaptive (or tailored) gamification~\cite{Klock2020TailoredLiterature}, rather than motivating users with solely external rewards, is designed to make the experience engaging for the specific user by also taking into account and adapting to individual traits according to necessary trade-offs. Nevertheless, adaptation withing a single gamification campaign might not be enough. People may still be uninterested regarding the activity promoted, or they might even be more useful for the community if actively engaged (also) in another initiative.

Given the issues described so far, we envision a new paradigm defined as \textbf{multi-challenge motivational systems}: by this paradigm, (micro-)communities participate to multiple campaigns devoted to common societal goals~\cite{DafflonGOD20}. Moreover, since participants' engagement owns a central and critical aspect, they form communities based on appropriate combinations of social capital contributed by each individual.

To develop multi-campaigns motivational systems, a new methodological paradigm and software platform is required. The paradigm combines multi-agent systems, gamification, and self-adaptive systems research field.  In fact, the systems should be able to: cater for groups and group phenomena such as smart communities (i.e. smart cities); facilitate and foster collective action; make citizen providers and consumers (prosumers) of collectively provisioned services; serve citizens in terms of social capital and needs, and make the \textit{collaboration of multiple citizens} the enabler to reach a common good, while detrimental (side) effects are detected and hopefully avoided.

The new methodological paradigm we propose is to ground the design of motivational digital  systems in theories of multi-agent systems in which agents are more likely to perform certain actions as based on motivational/engagement factors (i.e. through gamification). Moreover, the new platform we propose is based on the idea of \textbf{``Continuous Engagement''} \cite{RICHARD201880}, and is founded on formalisation of computational models derived from empirical analysis of psychological processes and social practices. The effects of different campaigns should be continuously evaluated to adapt the ongoing challenges to societal fluctuations and upcoming goals (hence, self-adaptive systems). The resulting platform provides the enablers for developing radically innovative \textbf{tools to motivate} in smart(er) communities and societies. We argue that the role of agents' capacity to support trade-off strategies should be capitalised~\cite{BencomoMoDRE2018}.

%
\section{Scientific Context}
\label{subsec:intro}

Multi-Agent Systems (MAS) address the process of how a community of agents with a joint interest work together in satisfying a high-level goal/objective in a specific context. The agents should be able to reflect on the current state of achievement of their goals, reason about how their actions might contribute to achieving their community-wide goal \cite{MeloMG18,MeirP18,PinheiroS18} and, how they can adapt the overall behaviours taking into account the preferences and the needs expressed by each single participant \cite{Bucchiarone19}. Research challenges in MAS cover the provision of models and techniques for \textbf{engagement}, \textbf{action}, \textbf{learning} and \textbf{adaptation} such that they perform and are interconnected as a community. This includes \textbf{micro-level modelling} pertaining to the individual actors, but also \textbf{macro-level modelling} of how
group practices emerge from composition of activities at the individual level. The representation of composition operators and how to apply them represent another open research challenge. We need mechanisms to handle how different individuals become part of the same emergent community, how they collectively devise and provide
services to the community, and how they collectively adapt to changes in context, interests, incentives and opportunities. If we want to facilitate and foster continuous engagement and collective actions of citizens using MAS, we need to understand how psychological and social processes can combine to pursue \textbf{self-organisation}. Dynamic Social Psychology (DSP) defines social groups as complex systems \cite{Vallacher97}, where the interaction between heterogeneous individuals or subgroups results in self-organisation and emergent properties at the system level. DSP incorporates social, psychological and cognitive mechanisms, which are empirically verified, in order to study how different psychological and social variables acting as control parameters have an impact the macro, group-level properties. 

Applying the DSP allows us to integrate findings on micro-level, individual determinants that have specific roles, capabilities and effects in social practices, as well as insights from system-level analysis, which can point to engagement policies that would optimise collective action to make it sustainable.
Based on motivational digital systems, citizens can share their “social capital” with the community. Social capital is an attribute of individuals that enhances their ability to improve a specific societal challenge taking actions in their daily life\cite{PetruzziBP15}.  These attributes may take different forms, for example reputation, participation, influence, support, among others. Each of these forms is also a subjective indicator (or metric) of one individual’s expectations about how other individuals will behave in an n -player cooperative context. In the pursuit of successful and sustainable actions, it is necessary to understand and capture the social practices that lead to the creation of social capital, the group psychology that provides an evaluation of social capital, and the mechanisms by which social capital contributes to reach the common good. We envision to build upon and extend gamification approaches, because they facilitate capturing sustainable practices in an algorithmic form, together with persuasion potentials to promote and incentivize the participation in those practices by citizens. A research challenge related to this aspect is \textbf{the automatic generation of personalised game experiences and mechanics} that are tailored to the user profile (e.g., preferences, habits, game status and history) and to the sustainability objectives promoted by the application. In fact, these techniques shall keep end-users engaged and interested in the long-term with a diversified and enhanced game experience, and at the same time incentivize sustainable behaviours .

There exists already a relevant body of research about modelling gamification applications \cite{KOIVISTO2019191}, in general they introduce processes to support a certain gamification solution, or they target a particular application domain (education, e-banking, health, etc.). However, the issues related to engagement and multi-objective games are scarcely addressed. Notably, Toda et al. \cite{TodaVI17} discuss in their survey major problems detected in gamified education and the most critical elements of the games construction. Consistently, the authors noted that the ranking approach and the game elements associated with it, shall be carefully designed to avoid counterproductive behaviours (e.g. performance degradation).

From a more technical perspective, modelling gamification mechanisms poses challenges to language design \cite{BucchiaroneCM20}. In fact, each game can be conceived as a set of rules through which each participant interacts with the environment: moves, actions, rewards, etc. are woven together to create an appropriate playground. If we consider the coordination of multiple gamified applications as proposed in our vision, then these rules are expected to grow even more. The recurrent solution of letting domain experts specify the game rules in requirement documents and then hard-code the rules in a corresponding application does not scale-up, especially if frequent adaptations are needed. On the other hand, giving domain experts the possibility of defining themselves game rules and generating automatically the corresponding application raises usability problems: diagrammatic solutions become quickly intractable with the number of variables involved in the game, while expressing constraints in logic formulas would be comparable to writing directly the implementation
code. From a broader perspective, the rendering of modelling concepts to the user is an open research problem, and this vision can contribute with appropriate solutions for the gamification domain.
In this regard, HCI (Human-Computer Interaction) researchers strongly argue for the benefits of adaptive experiences in gamification~\cite{Orji2017TowardsSystems}. Delivering ad-hoc content to players enhances engagement, and can contribute in the fulfilment of the system’s underlying goal~\cite{Kaptein2012AdaptiveSnacking}. Following hard-coded approaches, on the other hand, can lead to neutral or even detrimental results~\cite{Aldenaini2020TrendsReview}. Although the proven advantages of tailored game content, many gameful applications still rely on the ``one-size-fits-all'' approach~\cite{Hamari2014a}. Even when some kind of adaptation is applied, scholars and practitioners mostly employ hard-coded (i.e., rule-based) adaptation using survey-based profiling~\cite{Klock2020TailoredLiterature}. Nevertheless, players’ behaviours and preferences can change overtime. As such, dynamic (run-time) adaptation approaches are preferable, which better adapt to unforeseen behavioural changes. Those changes are even more likely to happen in persuasive gameful systems, where modifying behaviours is the goal. Towards this, players’ in-game activity and interaction patters can be exploited~\cite{Hooshyar2018Data-drivenReview}. While data has been used to learn specific preferences~\cite{loria2020reading}, adjust difficulty~\cite{khoshkangini2020automatic}, or provide recommendations~\cite{toda2019planning}, those studies tackled specific adaptation problems, often context-dependent. A tentative of a more elaborated adaptation framework was done by Loria~\cite{loria2019framework}. Yet, the framework presented is conceptual, and thus concrete solutions are still needed. Additionally, the framework is designed to adapt a single gameful application. Therefore, an approach to integrate knowledge on players' experiences deriving from several gamification is still lacking.

\section{Our Vision}
\label{sec:vision}

\begin{figure*}[h!]
	\centering
	\fbox{\includegraphics[width=.6\linewidth]{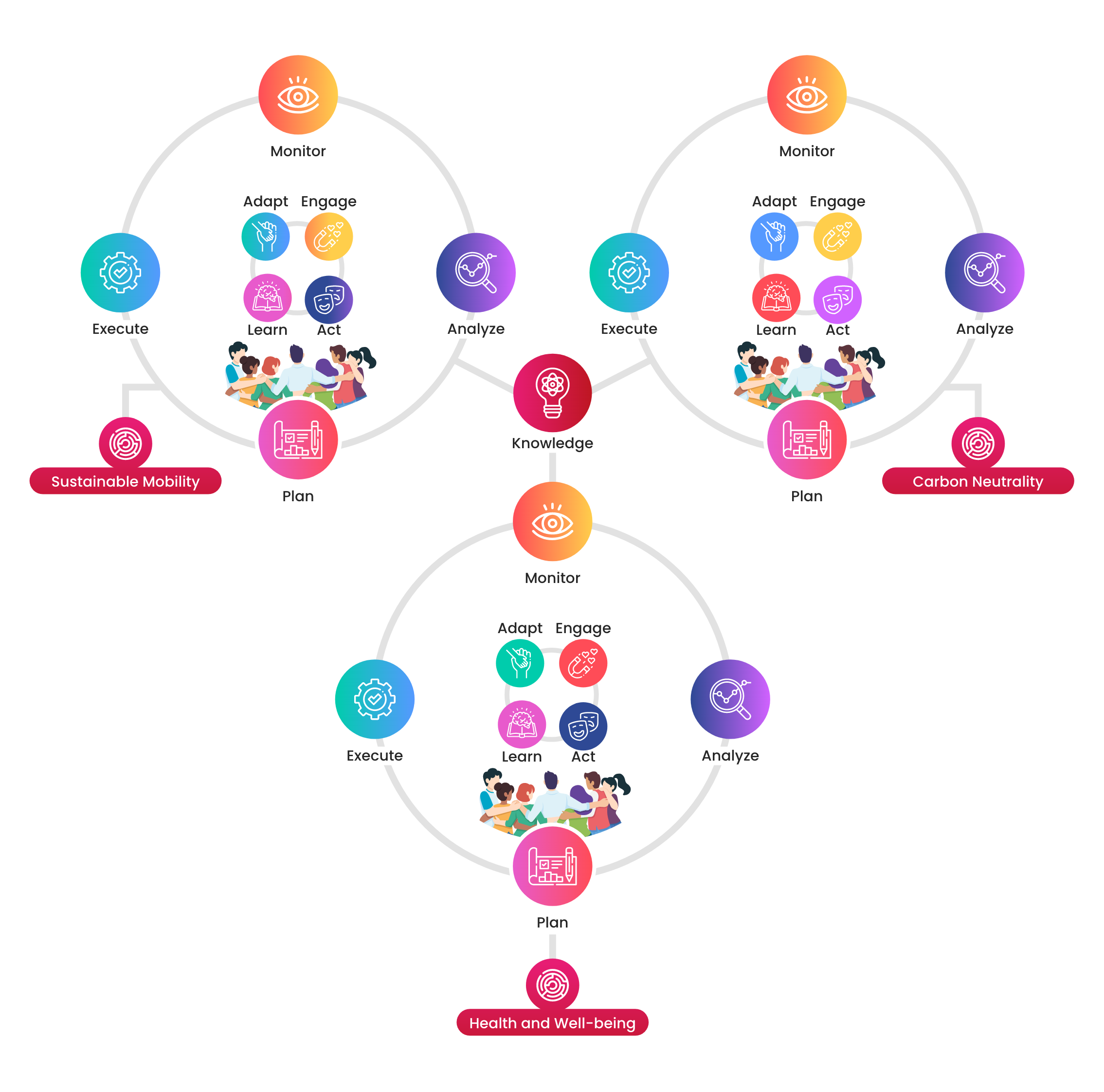}}
	\caption{The Role of Multi-Agent Systems in Motivational Digital Systems.}
	\label{fig:vision}
\end{figure*}
Our vision (depicted in Figure \ref{fig:vision}) is of a future where the "value'' of people will be based on their \textit{“contribution” to the society}. A society where many people will not have a job, instead they will voluntarily offer services to other people because they can and they like doing that (see as an example Uber, AirBnB, etc.). Other contributions to society can be having low impact behaviours (e.g. avoiding personal transportation, eating more vegetarian food due to lower carbon emissions, etc.), having healthy habits (exercise is not only a personal matter, active people have a lower cost on healthcare and have less ageing-related problems), sharing resources with others (not only cars). In short, the tendency is that society \textit{would shift from competition to cooperation}.

As depicted in Figure \ref{fig:vision}, our vision is based on a societal challenges-based approach that bring together resources and needs (i.e.,  \textbf{knowledge}) across different challenges (i.e., \textit{Sustainable Mobility, Carbon Neutrality, Health and Well-being}). This knowledge  reflects the policy priorities of each city/country strategy and addresses major concerns shared by citizens in their own communities. All activities done by a wide community (i.e. a city or a country) in a specific challenge shall take a \textit{continuous feedback loop} approach with the main purpose of improving the society, identifying specific initiatives capable to solve the contradictions of current societal problems and aiming at having a positive impact on the quality of life of citizens. To realise this, we exploit the powerful concepts and methods defined in the multi-agent systems research field.  In particular we exploit  the MAPE-K loop, the most influential reference control model for autonomic and self-adaptive systems \cite{Kephart2003}.  We envision a motivation digital system as a Multi-Agent System able to support the execution of the following flows: a \textit{top-down} flow able to manage from the city government to the citizens, and a \textit{bottom-up} flow that cover the emergent and evolving behaviours from the citizens to the overall city. The motivational digital systems implementing these two flows are composed by agents that see a smaller community, for example a city, inside a more general community, such as the society, from two different perspectives: the city and the citizenship. The first flow is composed by the following processes:

\begin{itemize}
    \item  \textbf{Monitor} is a process devoted to  ensure more effective monitoring and decision capabilities in cities exploiting local raw data about the specific societal challenge collected from heterogeneous sources (citizens, sensors, devices, and other city assets) to enable a pervasive, multi-source and multi-level analysis of the city and \textit{to help administrators, companies and citizens understand their city and how it evolves}.
    \item \textbf{Analyse} takes the results of monitoring, digesting them according to the challenge needs, and providing a set of  “alert” values spotting the relevant events that should trigger some adaptation; typically it includes mechanisms such as information fusion, complex correlation of information, identification of patterns, etc. For instance, sentiment analysis of twitter feeds could \textit{allow designers to see what people currently think and say about a specific societal challenge}. 
    \item \textbf{Plan} is a process taking the results of analysis, crossing it with available adaptation resources, and providing a set of plans to subgroups of citizen (micro-communities), in terms of a sets of plans; it typically includes mechanisms such as network partitioning, distributed consensus, local deliberation of plans, etc.
    \item \textbf{Execute} is a process fed with sets of plans, and able to interpret each as an aggregate process to be carried on by the associated communities of citizens; typically it includes all actuation mechanisms as for example \textit{citizen engagement and incentives for behaviour change}.
\end{itemize}
As soon as the various plans identified at the city level have been identified and executed, they have an effect on the city micro-communities.
For the second flow, we envision a set of  \textit{internal feedback loops} that involve different micro-communities in each specific societal challenge. Each loop is composed by: 

\begin{itemize}
    \item \textbf{Engage}. 
    The key objective of this phase is encouraging citizens to significantly change their daily habits, making them able to make more environmentally friendly and conscious choices. This can be done exploiting methods and tools (i.e., co-design) to bring the voice of the people into the discussion as part of the democratic process \cite{codesign}, or exploiting the motivational and persuasive power of games (gamification techniques) \cite{XI2019210}.
    
    \item \textbf{Act}. A main focus in such a motivational digital systems is to foster the ability and opportunity of citizens to self-organise in the form of collectively provided services (like in the sharing economy), through which citizens can act offering their expertise and facilitate the exchange, diffusion and adoption of virtuous practices by other fellow citizens. Another important focus is the exploration of gamification, as a way to incentivize and boost virtuous behaviours in a collaborative/competitive fashion. 

    This phase is introduced to provide innovative mechanisms computer-supported collective action in smart(er) communities and societies \cite{CA2014}.
    \item \textbf{Learn}. The goal of this process is to learn about the impact of the actions done by the different micro-communities in order to have a real time image of the city sustainability considering the different societal challenge (e.g., sustainable mobility, carbon neutrality, health and well-being). This is necessary to understand the health status of city-patient, its metabolism. It is possible thanks to new technologies \cite{CitySensing} and the attention paid by the \textit{Local Administration} to them.
    
    \item \textbf{Adapt}. The outcomes provided by the learning process will show the pros and cons of each city, the \textit{critical issues} and the strong points of the actions executed by their citizens. Based on this, the aim of this process is to adapt the engagement strategies and to define new challenges for their citizens. These adaptations will be useful for generating challenges for the micro-communities, taking into account both the city's objectives and the citizens' past performances and skills \cite{khoshkangini_2020}, that if realised will help to solve the problems highlighted.

\end{itemize}

A key aspect that characterises our approach is the provision of a unified way \textit{to manage different societal challenges as a combination of inter-operating motivational applications}. As a matter of fact, the existing research efforts address one societal challenge at a time (i.e., smart mobility, green, health, etc..); on the contrary, we believe that multiple challenges should be tackled simultaneously to maximise citizens potentials and benefits.

\section{Open Research Challenges}
The vision presented in this article introduces several research challenges, which are discussed next.

\begin{itemize}[leftmargin=*]
\item \textbf{Inclusive and Supportive engagement.} Our research vision includes the realization of an engagement framework sensitive to personal profiles. Profiling promotes \textit{inclusive} and \textit{supportive} engagement with citizens with diverse profiles and backgrounds (e.g. age, demographics, education, gender, disability, etc.), in order to fully understand the impact of diversity. 
Based on non-private  info associated profiles, our vision allows for creating a positive snowball effect in society, by leveraging the most motivated or active groups in society to progressively involve all the segments of local communities and help make first-person participation, individual commitment and citizen science more appealing, visible, and, eventually, conducive to change. To nurture citizens’ motivation to volunteer and act, specific models and tools are used to sustain the extended aspects of volunteering and its impact on society. These include \textit{Citizen Science} and technology-based \textit{crowdsensing} envisage the participation of the general public in societal challenges using an open and inclusive approach \cite{irwin_citizen_2002,10.11457}.  Motivational systems that properly exploit and embed game concepts and elements (gamification~\cite{DeterdingDKN11})  can support an engagement framework sensitive to diverse user profiles and backgrounds. Different types of incentives (e.g., monetary vs. non-monetary) and nudges are instead based on the fact that citizens tend to be strongly influenced by their family, friends and peers or (local) champions, and even wider through network effects\cite{canton_2014}, as such they should be used. \textit{Communities of practice}, \textit{living labs}, and \textit{crowdfunding} are further examples of the different approaches to citizen engagement that can be exploited \cite{0028914}.

\item \textbf{Adaptive and Continuous Engagement.}  The advertised value of gamification is the ability to turn unpleasant tasks into entertaining ones and improve users' experiences by fostering internal motivation to perform the task~\cite{Morschheuser2019TheCrowdsourcing}. Although successful gamification examples do exist, positive outcomes cannot be always assumed, as the effects vary with the contexts and implementations~\cite{hamari2014does}. The biggest critique to unsuccessful gamification applications is the ``one-size-fits-all'' approach, in which a static set of gamification elements are implemented (e.g., points, badges or leaderboards). Rather, researchers argue for the acknowledgement of players' diversity~\cite{Klock2020TailoredLiterature}. Studying gamification's effects have proved that gamification needs to be adaptive to reach its full potential in several domains (e.g., health, sport, and learning). Interpersonal differences contribute to the perception of game elements, as users are motivated by different elements~\cite{ryan2006motivational}. Consequently, designers and practitioners should prioritize the diversifications of the motivational affordances implemented towards the development of more inclusive gameful experiences~\cite{KOIVISTO2019191}. For instance, as different people are motivated by different things, personalizing the incentives and rewards can be highly beneficial~\cite{vassileva2012motivating}. As a consequence, tailoring at the level of social influence strategies may contribute in a positive way to the effects of persuasive technologies~\cite{Kaptein2012AdaptiveSnacking}. When individuals' preferences are neglected, in static gamification, researchers observed inconclusive or even negative results~\cite{Aldenaini2020TrendsReview,hamari2014does}. Thus, how to adapt and personalize gamified applications became an essential research area.

Many adaptation strategies rely on static adaptation (i.e. hard-coded or rule-based adaptations), using survey-based profiling approaches grounded in the Psychology literature. Although techniques aimed at understanding players’ preferences exist, adaptive gamification is still an under-investigated topic, especially for what concerns dynamic, automatic adaptation~\cite{Hallifax2019AdaptiveDevelopments,Klock2020TailoredLiterature}. Towards this, telemetry datalogs can be precious for learning players' specific preferences~\cite{loria2020reading}, adjust difficulty, and predict churn~\cite{loria2019exploiting}. 
Nevertheless, adaptive gamification research still misses consolidated research models and theoretical foundations~\cite{KOIVISTO2019191}. Researchers should further exploit implicit telemetry data ~\cite{Heilbrunn2017GamificationDesigns} to cyclically enhance the adaptation model~\cite{Klock2020TailoredLiterature}, leading to the periodical generation of novel content~\cite{KOIVISTO2019191}. While an initial tentative in the definition of a conceptual adaptation framework exist~\cite{loria2019framework}, this is only a step towards the right direction and still lacks a concrete, implemented solution.

\item \textbf{Evaluate the positive behavioural change.} A problem of gamification as a persuasive technology derives from the difficulty in ensuring a behavioural change. Players can either omit communicating activities misaligned with the system’s goal or the the application may even allow a partial recording of players’ actions. Consequently, a global view on players’ habits is lacking. For instance, if we consider an application for sustainable mobility pushing users to reduce car movements in favour of greener transportation means, we can measure a usage increase but cannot argue for players’ reduced car trips. In fact, an increase in green mobility does not imply less kilometres with other means. Our suggested framework also aims at moving forward in the solution direction. In the context of smart cities, 
knowledge retrieved from several gamification campaigns can be integrated to expand the understanding on players’ activity and (positive/negative) contribution to the society. Nevertheless, this is still a complex and challenging task. Researchers and practitioners need to develop algorithms to integrate this knowledge, while also avoiding  bias.

Positively transforming the way people behave for the benefit of society requires a deep transformation of their habits and behaviours, which must be based on comprehensive impact assessments and simulations that take into account social, health, environmental and climate impacts, as well as economic impact. This challenge aims to propose innovative solutions that provides data analytics, information, recommendations and simulations for city managers to assess, also through what-if analysis, the environmental impact, to evaluate changes as a result of specific measures and actions and to plan optimal and sustainable strategies. The global pandemic COVID-19 is an excellent example for this context.
\\

\item \textbf{Privacy and Ethical Issues.} 
The applications targeted provide collective behaviour by making use of shared knowledge (as depicted in Figure \ref{fig:vision}), and profile-based data as explained above. Privacy and ethical issues, therefore, have to become  first-class citizens when designing and evaluating this kind of systems. Hartswood \textit{et al.}~\cite{Hartswood:2015} discuss these issues in the context of peer profiling. They propose to return the control of data to users, supporting the principles of privacy and transparency. Other privacy-preserving mechanisms based on decentralized data ownership are proposed in the literature as a way to enhance privacy~\cite{Pournaras:2016}. However, in a world dominated by centralized platforms making profits out of the users' personal data, designers face a major socio-technical challenge in executing such a paradigm shift. The centralized/decentralized dichotomy affects the scalability of the proposed solutions too, of course, and so centralization may contain within itself the seeds of a move towards more decentralized solutions on technical grounds~\cite{deLemos:2013}.
A lot of work has been done at the European level on ethical and privacy guidelines for technology development with respect to this topic\footnote{\url{https://tinyurl.com/y36sq92q}}. We argue that by having a formal and explicit approach to define engagement policies can serve better as a means to define citizens rights in these contexts (in the same way it is done for e.g. in law). As part of our vision,  engagement mechanisms should intrinsically support diversity, ethics, etc. The feedback supported by MAPEK-loops is relevant to underpin the above, including  approaches to protect users by mediating their interactions with the digital world according to their own sense of ethics about actions and privacy of data \cite{Inverardi19}.

\end{itemize}

\section{Conclusion}
We believe that the vision presented here provides relevant steps towards the democratization of digital platforms for social coordination. Clearly, such a kind of research challenges can only be successfully tackled if addressed from a combined socio-technical perspective and if performed as an inherently multi-disciplinary research effort. Indeed, the practical implementation of motivational mechanisms needs continuous feedback from its theoretical counterpart to be found in sociology and psychology. The SEAMS community could play a potential relevant role in this multi-disciplinary undertaking.  Therefore, we have identified the following set of interrelated software engineering issues in his ambitious vision, which we hope are picked up by the SEAMS community: 

\begin{itemize}
\item The introduction of  a \textit{generic, multi-agent based, modelling methodology} to specify motivational digital systems in terms of targeted persons/communities and expected improvements.
\item Propose appropriate modelling concepts to define the interactions between different motivational digital systems, to avoid conflicting scenarios~\cite{BencomoMoDRE2018} and/or abuse, to respect privacy and take care of ethics .
\item The vision involves negotiation and coordination from different parties, which might lead to conflicting interests (e.g., widespread technology that promotes well-being can be a trade-off with low carbon emission, sustainability or financial return). This is closely related to requirements negotiation~\cite{SALAMA2017249} and the previous issue.
\item  The introduction of  mechanisms to handle the way how different individuals become part of the same emergent community, how they collectively devise and provide services to the community, and how they collectively adapt to changes in policies, context, incentives and opportunities.
\item Develop mechanisms to simulate, automatically derive and adapt multifaceted motivational means.
\end{itemize}

\section*{Acknowledgement}
This work was partially funded by the AIR BREAK project (UIA05-177), the Swedish Knowledge Foundation (KKS) Synergy project SACSys, the UK Lerverhulme Trust Fellowship "QuantUn" (RF-2019-548/9), and the UK EPSRC Project Twenty20Insight (EP/T017627/1).

\bibliographystyle{unsrt}
\bibliography{bib}

\end{document}